\documentclass{article}
\usepackage[utf8]{inputenc}
\usepackage{amsmath,amssymb,amsfonts}
\usepackage{algorithm}
\usepackage{algpseudocode}  
\usepackage{amsmath}  
\usepackage{color,xcolor}

\usepackage{graphicx}
\usepackage{textcomp}
\usepackage{fancyhdr}

\usepackage{tikz}
\usepackage{multirow}

\title{DNN Training Acceleration via Exploring GPGPU Friendly Sparsity}
\author{Zhuoran Song, Yihong Xu, Han Li, Naifeng Jing, Xiaoyao Liang, Li Jiang}
\date{Shanghai Jiao Tong University}

\begin{document}
\maketitle
\begin{abstract}
The training phases of Deep neural network~(DNN) consumes enormous processing time and energy. Compression techniques utilizing the sparsity of DNNs can effectively accelerate the inference phase of DNNs. However, it is hardly used in the training phase because the training phase involves dense matrix-multiplication using General-Purpose Computation on Graphics Processors (GPGPU), which endorse the regular and structural data layout.
In this paper, we first propose the Approximate Random Dropout that replaces the conventional random dropout of neurons and synapses with a regular and online generated row-based or tile-based dropout patterns to eliminate the unnecessary computation and data access for the multilayer perceptron~(MLP) and long short-term memory~(LSTM). We then develop a SGD-based Search Algorithm that produces the distribution of row-based or tile-based dropout patterns to compensate for the potential accuracy loss. Moreover, aiming at the convolution neural network~(CNN) training acceleration, we first explore the importance and sensitivity of input feature maps; and then propose the sensitivity-aware dropout method to dynamically drop the input feature maps based on their sensitivity so as to achieve greater forward and backward training acceleration while reserving better NN accuracy. To facilitate DNN programming, we build a DNN training computation framework that unifies the proposed techniques in the software stack. As a result, the GPGPU only needs to support the basic operator---matrix multiplication and can achieve significant performance improvement regardless of DNN model. Experiments results on MLP and LSTM using well-known benchmarks show that the speedup rate brought by the proposed Approximate Random Dropout ranges from 1.18-2.16~(1.24-1.85) when dropout rate is $0.3$-$0.7$ on MLP (LSTM) with negligible accuracy drop. As for CNN, the proposed sensitivity-aware dropout method can achieve up to $2.17\times$ speedup with $0.2\%$ accuracy loss. Our codes are available at https://github.com/songzhuoran/video-block-based-acc.
\end{abstract}


\section{Introduction}\vspace{-5pt}
Deep Neural Networks~(DNNs) have emerged as critical technologies to solve various complicated problems~\cite{Luo2017Understanding, Young2017Recent}. The inference of DNNs is computationally expensive and memory intensive and therefore has an urgent need for acceleration before we can fully embrace DNNs in the power-limited devices. Extensive works are proposed to reduce the computation by compressing the size of synaptic weights, such as weight pruning~\cite{Han2016Deep,Ding2017CirCNN}, quantization, low rank and Compact Network Design~\cite{Zhou2017Incremental,Gong2014Compressing,Leng2017Extremely}. 
The above compression techniques may require retraining the DNN with limited accuracy loss~($<1\%$). 
The success of these techniques relies on the sparsity and plasticity of DNNs; however, these techniques cannot directly apply to the training phase of DNNs.

The training phase, involving the back-propagation through the network to update the weights, demands three-times more computation effort. GPU is suitable for such task attributed to GPU's superior parallelism for large matrix multiplication~\cite{C2009Training, Puri2010Training}. Extensive works propose to accelerate the training phase on the distributed GPU-based system~\cite{Wen2017TernGrad,Zhang2016Staleness}.
Other works focus on accelerating the training phase using gradient pruning and weight quantization, respectively.

Random Dropout technique addresses the over-fitting problem and is widely used in training MLP and LSTM. The most common method randomly dropping some neurons of each layer in every training iteration, while the other (DropConnect~\cite{Wan2013Regularization}) aims the same goal by randomly dropping some synapses connections between layers~(i.e., some elements in weight matrix).
Theoretically, we can reduce the number of multiplication to $30\%$-$70\%$ if we can skip the calculation of all the dropped neurons or synapses while the dropout rate changes from 0.3 to 0.7.
However, the potential tremendous saving of multiplication and data access is hard to exploit because the neurons or synapses are randomly and irregularly dropped following the Bernoulli distribution. Such irregularity prevents the GPU's single instruction multiple threads~(SIMT) architecture to skip the unnecessary multiplication and memory access.

Therefore, for accelerating the training process of MLP and LSTM, we propose the approximate random dropout technique that replaces the random dropout with two types of regular dropout patterns to make the choices of dropped neurons or synapses predictable, which allow GPU to skip calculation of those dropped neurons or synapses. We further developed an SGD-based Search Algorithm to produce the distribution $\mathcal K$ of dropout patterns such that the dropout rate of each neuron is approximately subjected to a Bernoulli distribution (We provide a brief proof). In each iteration, we sample a dropout pattern subjected to $\mathcal K$ and then eliminate the redundant computation by omitting the dropped data during the hardware allocation. Consequently, the training process of MLP and LSTM can be substantially accelerated by our proposed regular dropout patterns.

Since the errors can accumulate and magnify through multiple layers, the previously published approximate random dropout technique will lead to severe accuracy loss if is directly implemented to the large CNN model. Dedicated training mechanism targeting for the CNN is therefore required to accelerate the CNN training process. Facing its increasing computing power and high training accuracy demand, 
we first explore the impact of the input feature maps and weights on the training accuracy to determine which one should we drop. And we find that the input feature maps have smaller influence than weights on the training accuracy. We consequently choose the input feature map as the drop target during CNN training. 

Since the input feature maps have different sensitivity as reported in the previous works, we propose the sensitivity-aware dropout technique that can capture the feature characteristics in the input feature maps during CNN training. The sensitivity-aware dropout technique first dynamically identifies the sensitive regions in the input feature map which are likely to contain important feature information for the final training results. It then applies low/high dropout possibilities to sensitive/insensitive regions. In this way, the characteristics of the input feature map will be reserved and fortified for higher CNN accuracy while saving a significant amount of computing resources from the insensitive regions. We further provide the dynamic dropout ratio tuning technique to dynamically tune the dropout ratio during the CNN training process to ensure accuracy. Moreover, considering the GPU's SIMT characteristic, we propose the hardware-aware dropout mechanism that drops the input feature maps in row- and block-level. 

Last but not least, to deploy the proposed techniques, we define the basic operator in them as the matrix multiplication. We then provide the DNN~\cite{Zhang2017ShuffleNet,Wen2017Learning,Zhou2017Incremental,Ding2017CirCNN} training computation framework that unifies the proposed techniques in the software stack. The computation framework first carries out the approximate random dropout and the sensitivity-aware dropout techniques to generate the binary mask for MLP, LSTM, or CNN. Afterward, the framework will constitute the matrix based on the binary mask that is used to indicate the keeping/dropping row/blocks. As a result, the GPU only needs to support the basic operator---matrix multiplication and can achieve satisfactory performance gain regardless of the DNN model.

Our experiments show that the speedup rate brought by the proposed Approximate Random Dropout ranges from 1.18-2.16~(1.24-1.85) when dropout rate is $0.3$-$0.7$ on MLP (LSTM) with less than $0.5\%$ accuracy loss. We find that when the batch size increases, the speedup rate increases with accuracy of neural network declines. As for CNN training acceleration, the proposed sensitivity-aware dropout method achieves $2.17\times$ speedup with negligible accuracy loss.

The reminder of the paper is organized as follows: Section~\ref{sect:related} introduces the background. Section~\ref{sect:motivation} illustrates the motivation of our proposed sensitivity-aware dropout design. Section~\ref{sect:dropout} describes the proposed Approximate Random Dropout Technique. Section~\ref{sect:non-algorithm} introduces the proposed sensitivity-aware dropout method. Experiments are shown in Section~\ref{sect:experiments}. Section~\ref{sect:conclusion} concludes this paper.

\section{Background}\label{sect:related}
\subsection{Accelerating DNN inference and training}
There are considerable works pitch into accelerating inference of DNN by leveraging the sparsity of DNN.
Han et al. prune synaptic weights which are close to zero and then retrain the DNN to maintain the classification accuracy. 
The zero weights are then indexed, compressed and moved onto the on-chip memory.
Special decoder is deployed in the accelerator to skip the computation of zero weights.
Consequently, above methods can only benefit ASIC/FPGA based DNN accelerator instead of GPU.
Jaderberg et al. and Ioannou et al. use low-rank representations to create computationally efficient neural networks. These methods cannot be used in training phase because of the subtle change of the weights degrades the convergence and accuracy of the training phase.


Extensive works propose to accelerate the training phase on the distributed GPU-based system~\cite{Wen2017TernGrad,Zhang2016Staleness}. Wen et al. propose to use ternary gradients to accelerate distributed deep learning in data parallelism. Zhang et al. propose a variant of the asynchronous SGD algorithm to guarantee the convergence of this algorithm and accelerate the training in a distributed system. 
Other works are relative to the acceleration in the training process using gradient pruning and weight quantization.
Kster et al~\cite{Han2016Deep}. share the exponent part of the binary coding of the weights and thereby convert floating-point operations to fixed-point integer operations. Noted that this work is compatible with ours and we leave this topic to further research.
Sun et al~\cite{Jaderberg2014Speeding}. prune those comparatively small gradients to speed up training phase. However, their work focuses on software-level optimization and thus yields marginal training acceleration while this work enable computation reduction on hardware-level.

\subsection{Basics of the GPGPU}\label{ssect:gpu}
GPGPU is commonly used for DNN training. 
It is composed of dozens of streaming multiprocessors~(SMs)~\cite{Wan2013Regularization,Wen2017Learning}.
Each SM consists of single instruction multiple threads~(SIMT) cores and a group of on-chip memories including register file, shared memory, L1D cache and etc.
Each SM manages and executes multi-threads on it. Those threads are clustered into warps, executing the same instruction at the same time. 
Thus, the branch divergence occurs when programmers write conditional branch (if-else).

Shared memory is a performance-critical on-chip memory. 
The latency of accessing the global memory (DRAM) is roughly 100x higher than that of accessing the shared memory. Hence, reducing the frequency of accessing global memory is critical for performance.
The capacity of the shared memory per block is 48KB in Nvidia GTX 1080Ti, which is much smaller than the capacity of the global memory. Therefore, reducing the superfluous data in shared memory is also important.

The key purpose of this work is to reduce the scale of matrices, by which we can reduce the access frequency of the shared memory and the global memory as well as the computation effort to accelerate the training.

\subsection{Random Dropout}\label{ssect:dropout}
Random dropout is widely used to prevent over-fitting. It randomly omits part of the neurons or synapses on each training iteration.
The probability of a neuron or a synapses to be dropped is subjected to a Bernoulli distribution parameterized with a \emph{dropout rate}~\cite{Srivastava2014Dropout}. In a nutshell, the main reason why random dropout can effectively prevent over-fitting is that it generates adequate different sub-models to learn diverse features during the training process and ensembles those sub-models to maximize the capability of DNN for inference~\cite{Wan2013Regularization}.




A question arises: why not skipping the calculation of those dropped neurons to reduce the redundant time spent on the matrix multiplication and the data movement.
Intuitively, we can write conditional branch~(if - else) to skip the redundant calculation. However, such conditional branches incur branch divergence in GPU, which is a great hurdle for performance.
In GPGPU's SIMT architecture, the red threads have to wait for the green threads. Thus, some process elements~(PEs) are idle, represented by the red cross. The total execution time is not reduced~(even increased) due to the branch divergence. Thus, it is non-trivial to exploit the dropout for speedup in GPGPU.

\section{Approximate Random Dropout}\label{sect:dropout}

\begin{figure}[tb]
\centering
\includegraphics[width=0.5\linewidth]{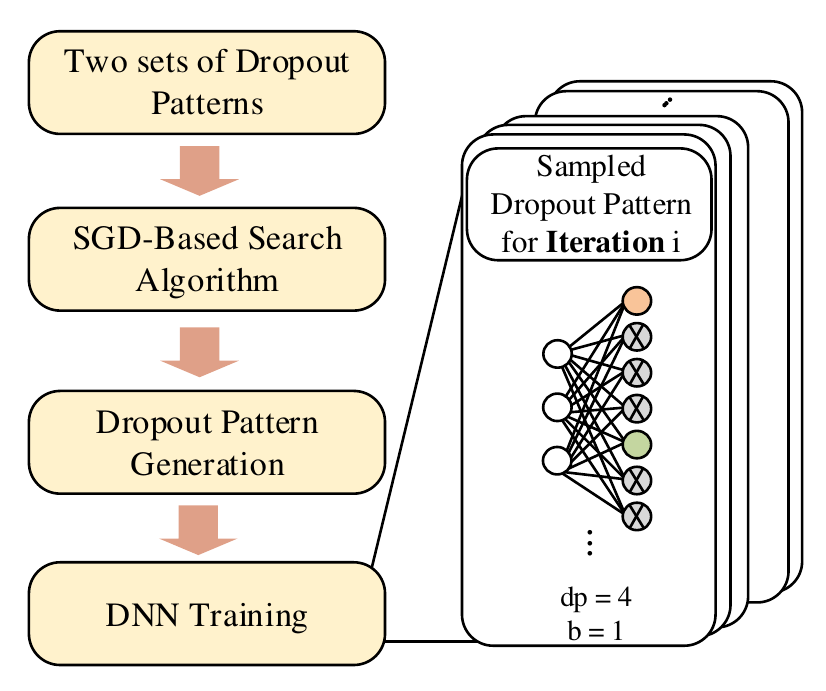}\vspace{-10pt}
\caption{Overview of the Approximate Random Drouput.}
\label{overview}
\end{figure}

\begin{figure*}[tb]
\centering
\includegraphics[width=0.7\linewidth]{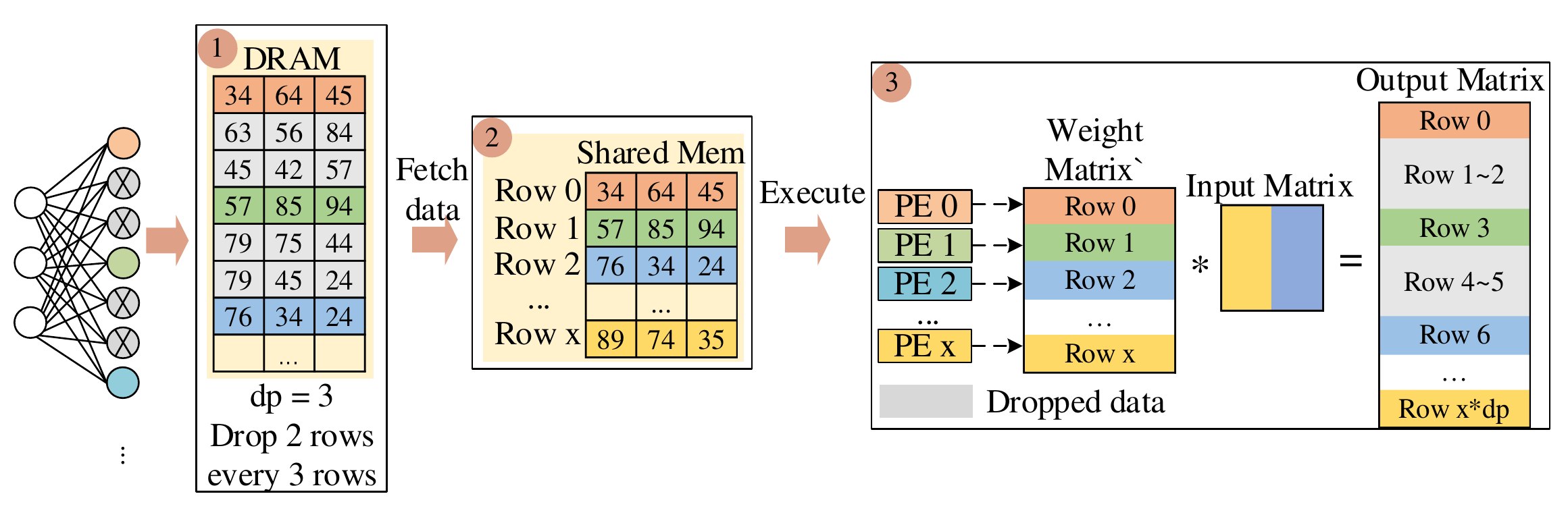}\vspace{-10pt}
\caption{Row-based Dropout Pattern.}
\label{pattern}
\end{figure*}

The key idea of accelerating the MLP and LSTM training is to reduce the scale of matrices involved in multiplication and avoid the divergence of GPU. However, the randomness in conventional dropout methods hamper the scale reduction.

For accelerating the MLP and LSTM, we define \emph{dropout pattern} as the combination of dropped neurons in each training iteration. As shown in Fig.~\ref{overview}, we replace the random dropout with regular dropout patterns generated online. Resulted from the replacement, we can forecast which neurons or synapses to be dropped and thereafter assist GPU to skip the calculation and data access of the dropped neurons without incurring the divergence. 

However, the loss of randomness induced by the regular dropout patterns increases the risk of over-fitting the MLP or LSTM.
To cope with this issue, we further develop a Stochastic Gradient Decedent~(SGD) based Search Algorithm~(see section~\ref{ssect:sgd}), to find a distribution of all possible dropout patterns such that the probability distributions of each neuron or synapse being dropped between our method and conventional method is equivalent. We provide a brief proof of that.


In this section, based on the computation characteristic of GPU, we firstly propose two sets of Dropout Patterns---Row-based Dropout Pattern (RDP) and Tile-based Dropout Pattern (TDP)---and then elaborate how to reduce computation and data access. After that, we introduce our SGD-based Search Algorithm which produce a distribution of possible dropout patterns as well as the dropout pattern generation procedure in each iteration.

\subsection{Row-based Dropout Pattern}

In conventional random dropout method, a dropped neuron incurs the multiplication of zero and the correspondent row in the weight matrix of next layer; in RDP, we drop the whole row in the weight matrix, which is equivalent to drop all the synapses of a dropped neuron.

Concretely, RDP is parameterized by two scalars $dp$ and bias $b$ as follow: we uniformly choose a bias $b\in \{1,..., dp\}$ and drop all rows in the weight matrix whose indices $i$ satisfy 
\begin{equation}
    i: (i-b) \bmod dp \neq 0
\end{equation}
Consequently, $(i-1)/(i)$ of the neurons are dropped. When $dp$=$3$, $b$=$1$, for instance (the left of Fig.~\ref{pattern}), we drop two rows~(i.e., neurons) in every successive three rows~(neurons) in the weight matrix from the top. Note that two scalars $dp$ and $b$ changes in every training iteration.

Given the size of the output matrix as $M \times N$, the maximum $dp$ is $dp_{max} = M$, and the maximum number of the sub-models is $\sum_{i=1}^{M} i=(M+1)/2$ considering the number of possible bias is $i$ when $dp=i$.

The execution processes in GPU is also shown in Fig.~\ref{pattern}. DRAM stores the whole weight matrix~(as shown in step 1); the gray block denotes the rows of weight matrix correspondent to the dropped neurons. We write the kernel function to prevent GPU from fetching those dropped data into shared memory~(as shown in step 2) and build two compact matrices (input matrix and weight matrix) for next step. After data fetch, every PE multiplies one row of the weight matrix by the whole input matrix. Thus, only $\frac{1}{dp}$ of the original weight matrix as well as the input matrix is fetched and calculated. The resulting rows fill $1\times dp$ rows in the Output Matrix using the same pattern. The rest $\frac{dp-1}{dp}$ of the Output Matrix is set to zero by default. Note that the RDP is agnostic to the matrix-multiplication algorithm as it temporarily compresses the matrices into a compact layout. Therefore, RDP can comply to any optimization method for matrix multiplication.

\begin{figure*}[tb]
\centering
\includegraphics[width=0.7\linewidth]{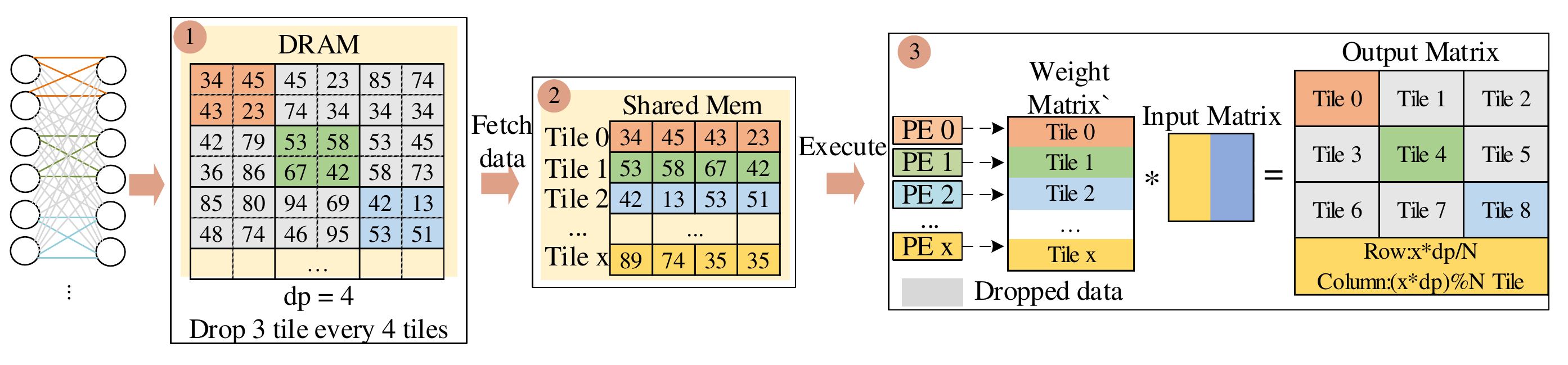}
\caption{Tile-based Dropout Pattern.}
\label{tile_pattern}
\end{figure*}

\subsection{Tile-based Dropout Pattern (TDP)}

Tile is a sub-matrix in weight matrix and contains multiple synapses connections. We use tiles as the unit to drop rather than synapse (namely the size of tiles is $1$) for the purpose of regularity. TDP is also parameterized by $dp$ and bias $b$. $dp-1$ tiles are dropped in every $dp$ tiles, resulting in $\frac{dp-1}{dp}$ of synapses connections being dropped.
When $dp=4, b=1$, as shown in the left of Fig.~\ref{tile_pattern}, we drop 3 tiles in every 4 successive tiles starting from first tile.

TDP has similar procedure compared to RDP but is different in two aspects: (1)TDP fetches non-dropped tiles into the shared memory rather than rows, and builds two compact matrices. (2) each PE conduct the multiplication of one tile of compact weight matrix and the corresponding tile of compact input matrix, according to their PE index. In the right of Fig.~\ref{tile_pattern}, GPU only conduct multiplication of two compact matrices whose scale is $\frac{1}{4}$ of the original scale.
This Dropout Pattern can naturally work with Tiling Method in matrix multiplication, which is an essential optimization technique.

Given the size of the output matrix $M \times N$, the size of the tile $x \times y$, the maximum $dp$ is $dp_{max} = \left \lfloor {M}/{x} \right \rfloor \times \left \lfloor {N}/{y} \right \rfloor$ and the maximum number of sub-models is $(1+dp_{max})/2$. 
TDP can generate more sub-models than RDP, when $N$ is roughly greater than $x\times y$.

The choice of tile size is critical: small tile leads to more diverse Dropout Patterns as well as sub-models, but fine-grained control.
Under such circumstances, the size of tile is set to be $32\times 32$ to balance the maximization of the number of sub-models and avoiding shared memory's bank conflict since the shared memory has 32 banks in NVIDIA GPU.

A typical training process of MLP and LSTM is composed of three steps: fully connected layer computation, activation layer computation and dropout layer computation using the mask matrix. 
After applying the Dropout Pattern with $dp = 2$, we only need to spend half of the time for fully connected layer computing and skip the dropout layer computing. 
Consequently, given the dropout pattern, the time spending on training can be overtly reduced.

\subsection{SGD-based Search Algorithm for Dropout Pattern Distribution}\label{ssect:sgd}

For each iteration in training procedure, only one regular dropout pattern is applied to the network. In order to approximate the traditional dropout process, the dropout pattern we choose in each iteration should satisfy that: (1)the dropout probability of each neuron should subject to a given Bernoulli distribution, and (2)different sub-models derived from that series of dropout patterns should be adequate.

With the above two requirements in mind, we propose an efficient SGD-based Search Algorithm to find a desired probability distribution. We then randomly generate the dropout patterns following the found probability distribution. Statistically, we want the derived dropout patterns can satisfy the above two demands. SGD consumes tractable time and is convenient in optimizing the continuous variables.
More specifically, the algorithm obtains a probability distribution $\mathcal K=\{k_i\}_{i=1}^{dp_{max}}$ which contains the probability $k_i$ of each possible Dropout Pattern $i\in \{1,2,..., dp_{max}\}$, which is subjected to $\sum_{i=1}^{dp_{max}}{k_i} = 1$. 

Here we define the \emph{global dropout rate} as the proportion of neurons or synapses who are set to zero. Noted that the global dropout rate is different from the conventional dropout rate which refer to the probability of a single neuron or synapse to be dropped. However, we prove that within our approach the two dropout rate are statistically equivalent.


Given the target global dropout rate $p$, and the maximum $dp$ as $N$, we use Algorithm~\ref{algo} to search for desired distribution $\mathcal K$.
A vector $v$ with length $N$ is first arbitrary initialized (line 1) and the $softmax(v)$ serve as the final probability distribution of each dropout pattern (line 4). Then we setup a constant vector $p_u = \{\frac{i-1}{i}\}_{i=1}^{N}$ whose element denotes the global dropout rate of a given dropout pattern. Consequently, $d^T\cdot p_u$ is the expected global dropout rate and the difference between it and the target global dropout rate is our optimization objective (line 5). The negative information entropy of $d$ is added to the loss to enforce $d$ to be dense and to produce more diversified sub-models (line 6, 7).
The algorithm uses SGD algorithm to update $v$ (line 8, 9) and return the distribution $d\in [0,1]^{N}$ when the loss is stuck.
The loss function defined in line 7 derives the algorithm to find a distribution $\mathcal K$ that (1)makes the global dropout rate equal to required value $p$ and (2)maximizes the sub-models diversity.

\subsection{Dropout Pattern Generation}

The acquired distribution $\mathcal K$ is then used to sample dropout pattern in each iteration. In each iteration, we randomly sample a dropout pattern (parameterized by $dp$ and $b$) subjected to the distribution $\mathcal K$, and then uniformly choose a bias $b\in \{1,..., dp\}$. Dropout pattern is then determined.

In our method, global dropout rate is statistically equivalent to the single neurons or synapse dropout rate. For each neuron or synapse, the probability of it to be dropped (conventional dropout rate) is:
\begin{equation}
p_{n}=\sum_{i=1}^{dp_{max}}p_b k_i=\sum_{i=1}^{dp_{max}}\frac{i-1}{i}k_i
\end{equation}
The global dropout rate of $\mathcal K$ is:
\begin{equation}
p_g = d^T\cdot p_u = \sum_{i=1}^{dp_{max}} k_i\frac{i-1}{i} \approx p
\end{equation}
Therefore, in terms of the whole training process, the dropout rate $p_n$ of a single neurons or synapse is equal to the global dropout rate $p_g$ and thus is approximately equal to the target dropout rate $p$ by the SGD-based Search Algorithm.

\section{Exploring the Sensitivity of Feature Map}\label{sect:motivation}

Based on the value distribution in the input feature maps of the CNN model, we observe that the majority of feature map values are close to zero, while a small number of feature map values are large. Because dropping values reduces the NN training accuracy as revealed by prior work, three questions naturally arise:
1) what type of data is more important during CNN training? weights or input feature maps? 2) whether there are input values in input feature maps that are crucial to the CNN training accuracy?; 3) how are these input values distributed in the spatial domain?

To answer the questions, we first find that input values are less important compared to weights so that we should drop input values instead of weights during CNN training. Moreover, we also observe that a majority of input values have smaller influence on the CNN accuracy and they aggregate in space that are termed as \emph{insensitive regions}. As a result, insensitive regions should be tracked and dropped in a hardware friendly manner to accelerate the CNN training process, while the sensitive regions should be reserved for negligible CNN accuracy loss. This motivates our feature map sensitivity-aware dropout algorithm as elaborated in Section~\ref{sect:non-algorithm}.

\subsection{The Importance of Weights}

\begin{figure}[tb]
\centering
\includegraphics[width=0.5\linewidth]{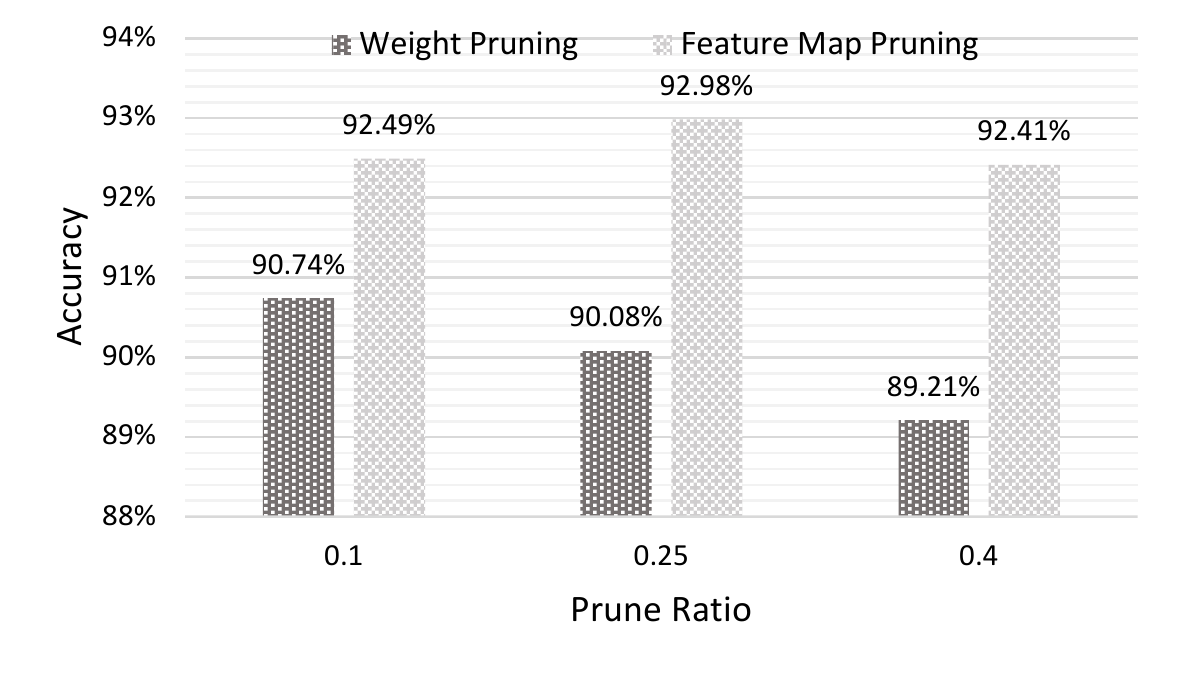}\vspace{-10pt}
\caption{Comparsion between pruning feature map and pruning weight.}
\label{fig-comp-w-i}
\end{figure}

In this section, we will explore the roles of input feature maps playing in the CNN training process so that we can decide which types of data should we choose to drop. We take MNIST dataset on AlexNet as a case study.

First, we drop weights and input feature maps separately. Then we measure the CNN accuracy to study the significance of weights and input feature maps. As shown in Fig.~\ref{fig-comp-w-i}, we find that dropping weights during the training process seriously deteriorates the CNN accuracy to $89.2\%$ when we drop $40\%$ weights. Alternatively, the CNN accuracy still remains $92.41\%$ when we drop $40\%$ input feature maps, as illustrated in Fig.~\ref{fig-comp-w-i}. We can conclude that the input feature maps have a smaller impact on the final CNN accuracy compared to weights. This is because dropping weights will make the training process becoming invalid and lead to a sharp accuracy degradation. As a result, we choose the input feature map as the dropout target to speed up the training process while reserving the CNN accuracy.

\subsection{Sensitivity of Input Feature Map Values}

\begin{figure}[tb]
\centering
\includegraphics[width=0.5\linewidth]{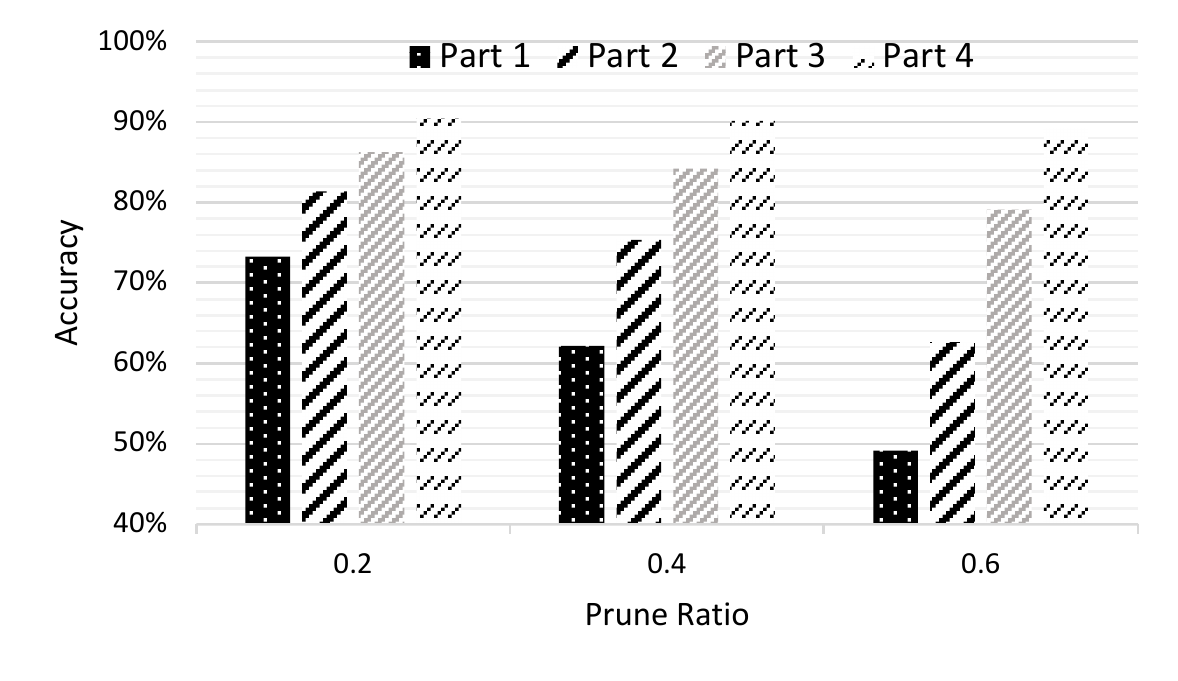}\vspace{-10pt}
\caption{Comparing the impact of different sensitive values on specific network.}
\label{fig-region-sensitive}
\end{figure}

In this section, we will investigate the sensitivity of input feature maps that affects the CNN training accuracy. We take the widely used dataset CIFAR-10 on VGG19 as case study. 

First, we classify the input feature map values of each layer into several parts according to their magnitudes at runtime. For example, we classify the values into four parts. Part 1 contains the largest $25\%$ values of the feature map, part 2 and 3 contain the middle $50\%$ values, and part 4 contains the smallest $25\%$ values. We then drop values of different parts with different dropout ratio during the training process. Finally, we measure the final CNN accuracy and study the sensitivity of different parts. Fig.~\ref{fig-region-sensitive} shows the results, where we have several bars representing accuracy results of different parts. For example, bar 1 represents dropping part 1, while bar 2 represents dropping part 2. And we can see that part 4 holding the smallest input values can maintain accuracy even when we drop $60\%$ values in part 4. In addition, the CNN accuracy decreases drastically when we drop values in part 1 no matter what the dropout ratio is. Those observations indicate that different input values in different parts have different influence on the CNN training accuracy that can be seen as the sensitivity of input feature maps. For example, part 1 containing the largest values is more sensitive to accuracy than other parts. In other words, we can guarantee the accuracy by carefully 
dropping the sensitive parts, while freely dropping the insensitive parts. But identifying the sensitive and insensitive values is still a big challenge as the input feature maps are not available until run time. This process needs to be efficient and hardware friendly, and therefore prompts our sensitivity-aware dropout technique as elaborated in Section~\ref{sect:non-algorithm}.

\subsection{Sensitivity of Feature Map Regions}

In this section, we will explore how the sensitive input values are distributed in the spatial domain. We take CIFAR-10 on VGG19 as an visualized example. We randomly peak one training iteration, and then use different colors to mark the input values into three parts as depicted in Fig~\ref{SensitiveRegion}. We can clearly see that the values with the same sensitivity gather together that terms as \textbf{sensitive regions}, while the insensitive values belonging to parts 2 and 3 dominate the input feature map.

As a result, we can apply the structured dropout pattern on the insensitive regions considering the SIMT characteristics of GPU. Thanks to the majority of insensitive regions, we can achieve tremendous speedup with the runtime structured dropout. Therefore, we propose the sensitivity-aware dropout scheme to accelerate CNN training performance with negligible accuracy loss, as illustrated in the following section.


\section{Sensitivity-aware Dropout Method}\label{sect:non-algorithm}

Section~\ref{sect:motivation} has shown that there are sensitive regions in the input feature maps during the training process. Based on this observation, we propose the Sensitivity-aware Dropout Method to speed up the CNN training process. There are three more problems that need to be addressed:
\begin{enumerate}
    \item How to design a sensitivity identification algorithm? As the input feature maps of CNN are dynamically changed, their sensitivities have to be identified efficiently and hardware-friendly.
    
    \item How to conduct the sensitivity-aware dropout with efficiency and high accuracy? For sensitive and insensitive regions, we should apply different dropout possibilities. As a result, we can reserve as many key features in sensitive regions as possible to maintain the training accuracy. In addition, the dropout granularity should be carefully considered. Compared to the fine-grained dropout, the coarse-grained dropout is more hardware-friendly considering the SIMT characteristics of GPU. Therefore, we will drop input feature maps in a coarse-grained manner.
    
    \item How to determine the dropout ratio? Given that the CNN accuracy varies during training, we should dynamically tune the dropout ratio so as to balance the final performance improvement and accuracy.
    
\end{enumerate}

To this end, we propose the sensitivity-aware dropout method mainly contains three steps: 1) sensitivity prediction; 2) hardware-aware dropout; and 3) dynamic dropout ratio tuning: 
First, we design a sensitivity prediction algorithm to efficiently locate the sensitive regions in the input matrix. Then, we generate a sensitivity mask, which records the dropout possibilities for sensitive and insensitive regions. We will provide more details in Section~\ref{ssect:sensitivity-pre}.
Afterward, we drop the input matrix in a hardware-friendly manner, which is elaborated in Section~\ref{ssect:hardware-aware-prune}. During the sensitivity-aware dropout, the dropout ratio will be dynamically tuned considering both speedup and accuracy. We will give the details in Section~\ref{ssect:dynamic}.


\subsection{Sensitivity Prediction}\label{ssect:sensitivity-pre}

Given input feature map of $C\times H \times W$ dimension and weight of $C_{out}\times C\times K\times K$ size, GPU first calls Im2col function to transform the input feature map to input matrix of $(H\times W) \times (C\times K \times K)$ size, and the weight to weight matrix of $C_{out} \times (C\times K \times K)$ size~(as shown in Step 1). Next, we split the input matrix into multiple $x\times y$ regions. For each region, we need to identify its sensitivity. For real-time sensitivity prediction without hurting the CNN training speed, we propose to randomly select $k\%$ values to determine the sensitivity of the region~(Step 2). We then compare $k\%$ values to the predefined threshold. The comparison process can be treated as using the step activation function. The corresponding region is sensitive if there exists $t\%$ values are larger than the threshold. Otherwise, the region is insensitive~(Step 3). Note that $k\%$, $t\%$ are empirically chosen by a subset of training dataset. Finally, we use different dropout possibilities for different regions, which will be recorded in the sensitivity mask with a dimension of $\frac{h \times w}{x\times y}$~(Step 4). Specifically, we apply low dropout possibility for the sensitive region, while high dropout possibility for the insensitive region. For instance, we apply $m\%$ dropout possibility for the sensitive region, while $n\%$ dropout possibility for the insensitive region~(Note that $n>m$). This can be applied to the following hardware-aware dropout technique in Section~\ref{ssect:hardware-aware-prune}. 

\begin{figure*}[!t]
\centering
\includegraphics[width=\linewidth]{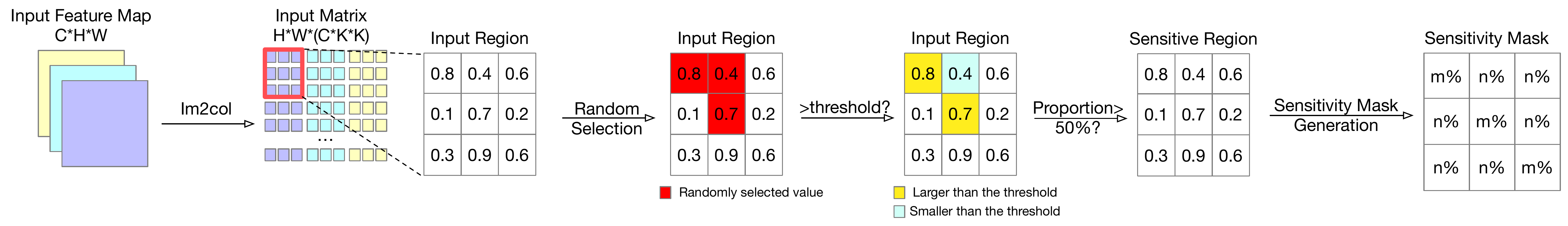}
\caption{Overview of Sensitivity Prediction.}
\label{fig-prediction}
\end{figure*}

Fig.~\ref{fig-prediction} shows the example of sensitivity prediction. To identify the sensitivity of one input region on-the-fly, we first randomly select $30\%$~($k\%$) values in the region and then compare them with the predefined threshold 0.5. Since the proportion of values that are greater than 0.5 is larger than $50\%$~($t\%$), we define the region as sensitive region. We iteratively execute the sensitivity prediction and generate the sensitivity mask indicating the dropout possibilities of sensitive and insensitive regions.


\begin{figure*}[!t]
\centering
\includegraphics[width=0.8\linewidth]{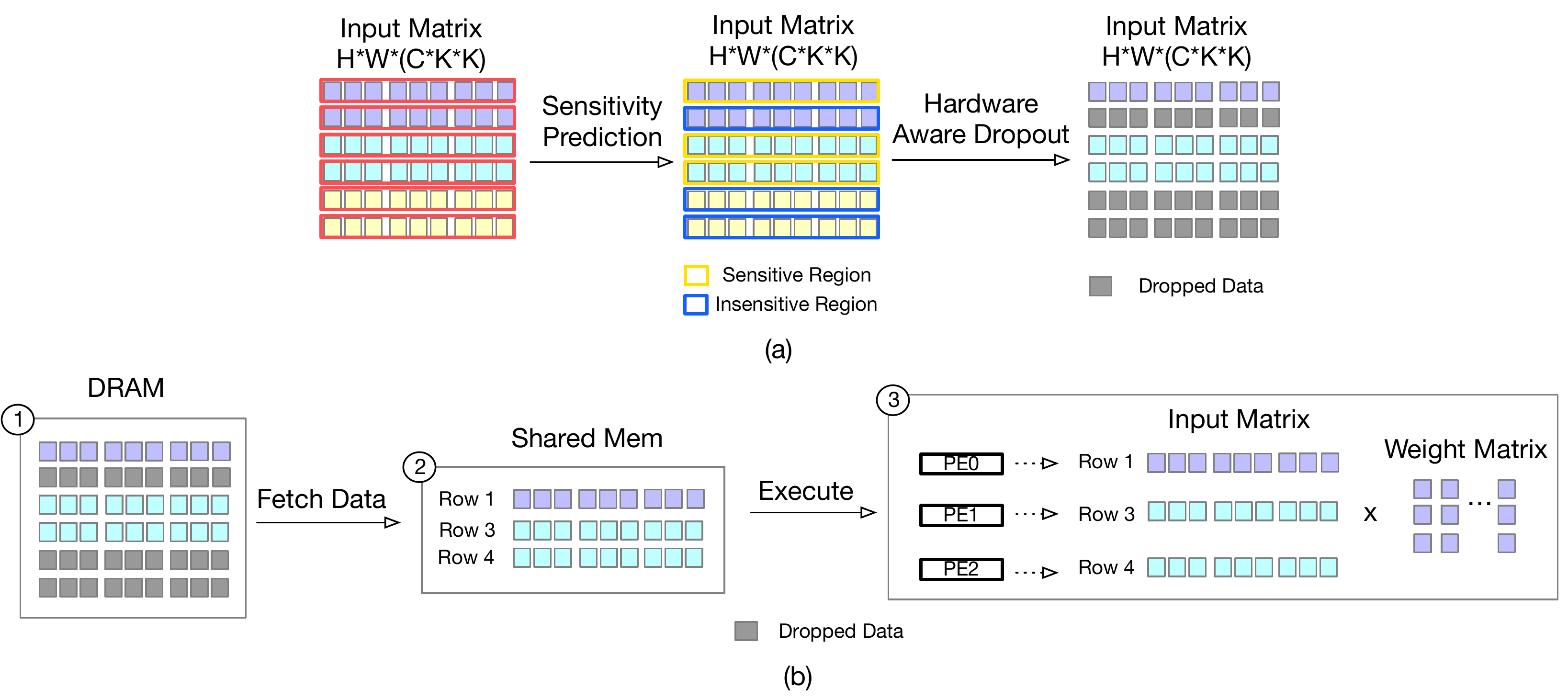}\vspace{-10pt}
\caption{Example of Row-based Sensitivity-aware Dropout.}
\label{fig-row-conv}
\end{figure*}

\subsection{Hardware Aware Dropout}\label{ssect:hardware-aware-prune}

To accelerate the CNN training process considering the GPU's SIMT characteristic, we drop the input matrix using the row-based or block-based sensitivity-aware dropout pattern.


\subsubsection{Row-based Sensitivity-aware Dropout Pattern}

In the Row-based Sensitivity-aware Dropout Pattern (RSDP), a row of input matrix can be seen as the dropout granularity. 
And we will drop the rows in the input matrix based on their corresponding dropout possibilities in each training iteration. We depict an example of RSDP in Fig.~\ref{fig-row-conv}(a). In this example, after sensitivity prediction, the first, third, and fourth rows are predicted as sensitive rows and they have $m\%$ possibility to be dropped. And since the second, fifth, and sixth rows are predicted as insensitive rows, they have $n\%$ possibility to be dropped. Next, we apply the hardware aware dropout to each iteration and drop a subset of rows in the input matrix. In the current iteration, the second, fifth, and sixth rows are dropped.

The execution process of the hardware aware dropout is shown in Fig.~\ref{fig-row-conv}(b). Initially, DRAM stores the whole input matrix~(as shown in Step 1). During each CNN training iteration, we will first sample the dropped rows in the input matrix based on their dropout possibilities. Then we write the kernel function to avoid fetching the dropped rows of input matrix into shared memory~(Step 2) and build the dense and small input matrix for next step. Afterward, each PE multiplies one row of the input matrix by the whole weight matrix~(Step 3). Next, the resulting rows fill rows in the output matrix using the same pattern. In short, as only a subset of the input matrix is fetched and calculated, the CNN training process can be well accelerated. 




\begin{figure*}[!tb]
\centering
\includegraphics[width=0.8\linewidth]{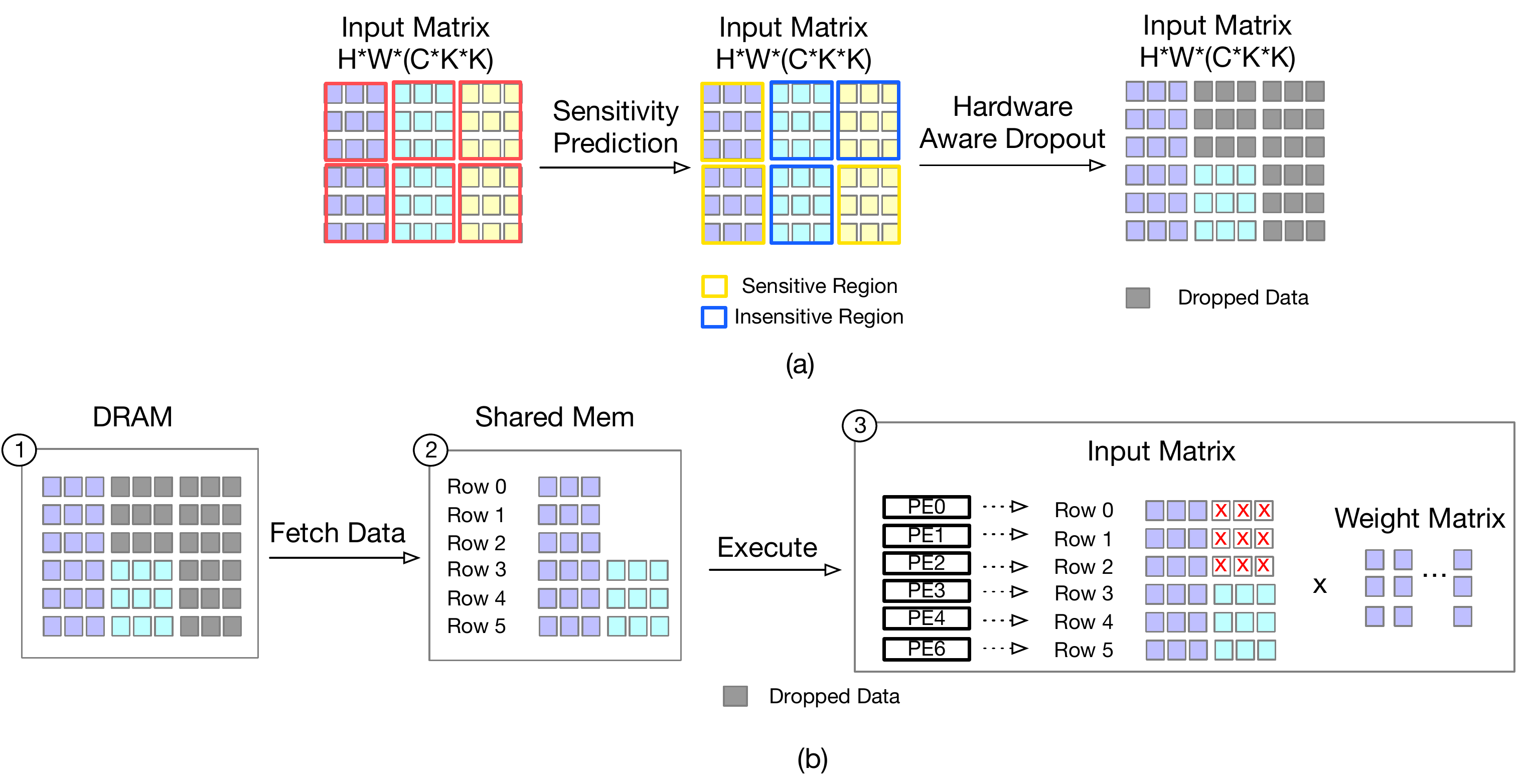}\vspace{-10pt}
\caption{Example of Block-based Sensitivity-aware Dropout.}
\label{fig-block-conv}
\end{figure*}

\subsubsection{Block-based Sensitivity-aware Dropout Pattern}

Given two large matrices, GPU applies tiling technique to optimize the matrix multiplication and memory accesses. Motivated by this, we propose the Block-based Sensitivity-aware Dropout Pattern (BSDP) that regards the block of input matrix as the dropout granularity.

Intuitively, we can directly drop the blocks in the input matrix at run-time. In this way, the second, third, and sixth blocks are sampled to be dropped, as depicted in Fig.~\ref{fig-block-conv}(a). However, given that the rows of the input matrix have uneven sparsity and each PE multiplies one row of the input matrix by the whole weight matrix, the PEs' workloads are unbalanced, as shown in Fig.~\ref{fig-block-conv}(b). In a nutshell, dropping the blocks by only considering the sensitivity results in lower utilization of PE and hampers the performance improvement.

To handle the extreme workload imbalance, we propose to redistribute the sensitive blocks and make them distributed evenly. Specifically, we first define the blocks that belong to the same row as a group. We then predict the sensitive blocks within each group. During the prediction, we will limit the number of sensitive blocks in all groups to be nearly equal. As a result, given that the rows of the input matrix have even sparsity, the PEs' workloads will be balanced. The detail of the block-based sensitivity-aware dropout considering the workload balance is shown in Fig.~\ref{fig-block-conv-new}.


During the conventional CNN training, there are two steps: convolution layer computation and activation layer computation. 
To implement our proposed method, we insert the sensitivity prediction and RSDP or BSDP operations before the convolution layer computation to determine which rows or blocks in the input matrix will be pruned. Consequently, we only need to spend little time for convolution layer computing and skip the pruned rows or blocks computing. Given the dropout pattern, the time spending on training can be greatly reduced.

\subsection{Dynamic Dropout Ratio Tuning}\label{ssect:dynamic}


To reduce the accuracy loss, we propose the Dynamic Possibility Pruning technique to dynamically tune pruning possibility during CNN training.


The input feature map is calculated by convolving the randomly initialized weight matrix and input data as the first step. Obviously, the initial input feature map is insufficient to extract key information in the input data. Therefore, pruning too many input feature maps at first will delay the incrementation of CNN training accuracy. As depicted in Fig.~\ref{dynamic1}, compared to the baseline, the accuracy can increase more when we use a lower prune possibility~($20\%$) given the training epochs. Therefore, we start with a low prune possibility to assist the CNN model extracting features. Afterward, we should gradually increase the prune possibility to achieve tremendous training speedup. However, to guarantee the final training accuracy reaches to a satisfactory degree, we use a relatively low prune possibility in the last period until the CNN model converges. 

In summary, we can see that the trend of the prune possibility obeys the skewness distribution that is shown in Eqn.~\ref{eqn-skewness}. And the remaining challenge is to generate the mean value $\mu$ and the standard deviation $\sigma$ in the skewness distribution so that we can obtain the prune possibility given a training epoch. As shown in Eqn.~\ref{eqn-ey}, the mean value $\mu$ can be calculated given the expectation $E(Y)$ as the total prune possibility. Moreover, the standard deviation $\sigma$ can be calculated using the similar way to that used for the mean value $\mu$, as shown in Eqn.~\ref{eqn-dy}. 

\begin{equation}\label{eqn-skewness}
    f_{Y}(y)=\frac{2}{\sigma}\varphi(\frac{y-\mu}{\sigma})\Phi(\lambda\frac{y-\mu}{\sigma})
\end{equation}

\begin{equation}\label{eqn-ey}
\begin{split}
    E(Y)&=\mu+\mu_{0}(\lambda)\sigma \\\\
    \quad \mu_{0}(\lambda)&=\sqrt{\frac{2}{\pi}}\frac{\lambda}{\sqrt{1+\lambda^{2}}}
\end{split}
\end{equation}

\begin{equation}\label{eqn-dy}
\begin{split}
    D(Y)&=\sigma_{0}^{2}(\lambda)\sigma^2 \\\\
    \quad \sigma_{0}^{2}(\lambda)&=1-\mu_{0}^{2}=1-\frac{2}{\pi}\frac{\lambda^{2}}{1+\lambda^{2}}
\end{split}
\end{equation}

\section{DNN Training Computation Framework}
To build the programmer-friendly interface and support the proposed methods, we propose the DNN training computation framework. 

\begin{figure}[!tb]
\centering
\includegraphics[width=0.8\linewidth]{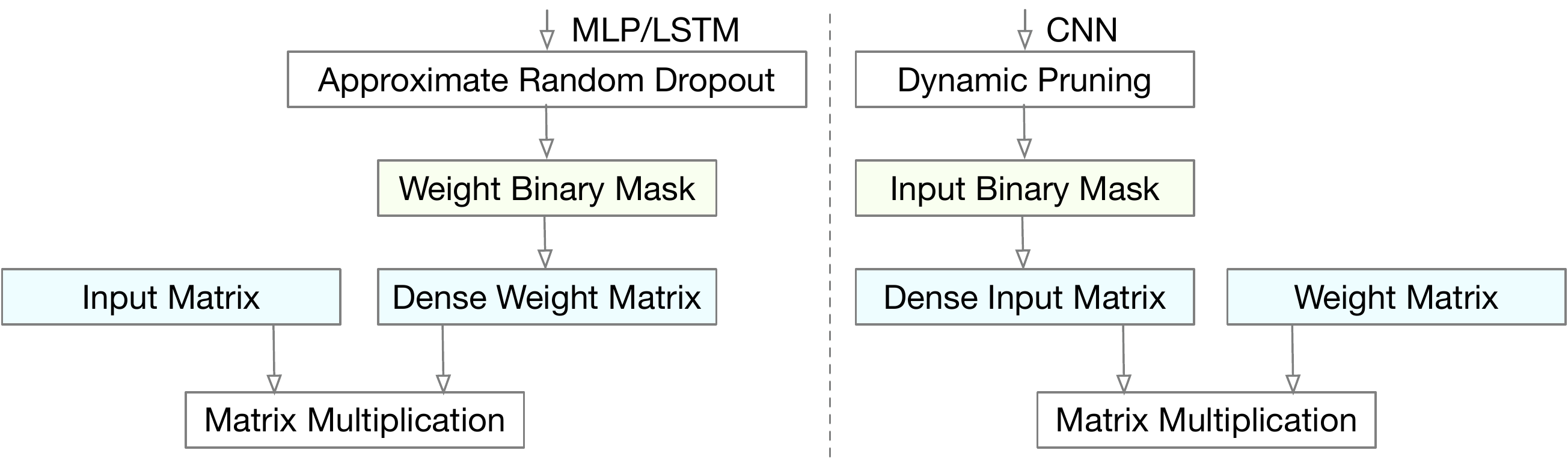} 
\caption{The overall DNN training flow using the proposed methods.}
\label{fig-framework-flow}
\end{figure}

The key ideas of the proposed approximate random dropout and dynamic pruning technique are both generating the sparse and structured weight or input matrix for eliminating the unnecessary computations during DNN training. The overall DNN training flow after applying the proposed methods is shown in Fig.~\ref{fig-framework-flow}. Specifically, for MLP and LSTM models, we leverage the weight binary mask to record the positions of zeros in the weight matrix. Moreover, we leverage the input binary mask to record the positions of zeros in the input matrix for the CNN model. For example, for a CNN model, given the input matrix that has six rows and we choose to skip the computations in row-level, we will generate the binary mask with six values, where ``0" indicates the row should be zero while ``1" indicates the row should be reserved. And the binary mask is ``1,0,1,1,0,0" for the example in Fig.~\ref{fig-row-conv}(a). Consequently, GPU can fetch the non-zero data from DRAM to the shared memory by utilizing the binary mask. Moreover, the PEs will execute the dense matrix multiplication and write the results back to their corresponding positions according to the binary mask. The detail execution process in GPU is depicted in Fig.~\ref{fig-row-conv}(b).

\begin{figure}[!tb]
\centering
\includegraphics[width=0.7\linewidth]{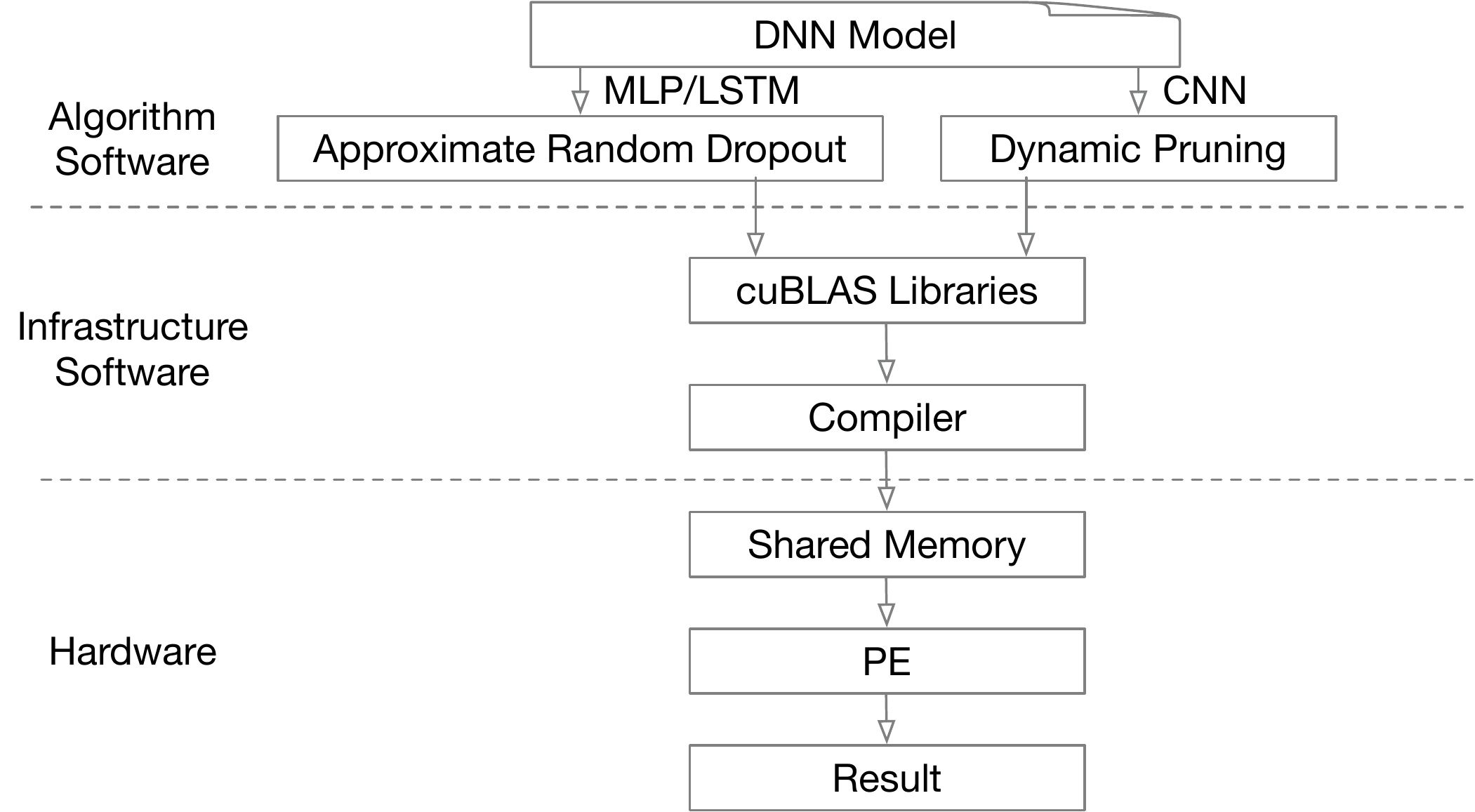} 
\caption{The DNN Training Computation Framework.}
\label{fig-framework}
\end{figure}

In the proposed computation framework, the programming flow is quite similar to the conventional DNN training flow, which is illustrated in Fig.~\ref{fig-framework}. The framework first invokes the approximate random dropout or the dynamic pruning technique for generating the binary mask given the DNN model. Then it launch kernels to constitute the dense matrix. In the infrastructure software, the framework applies the highly optimized CUDA library such as cuBLAS or cuDNN to efficiently execute the matrix multiplication. And in the hardware, the shared memory will only fetch the non-zero values from DRAM, which significantly reduces the memory stress. Moreover, the PEs only consume the non-zero values and conduct the dense matrix multiplications.

In summary, our proposed methods can be well adopted in the proposed DNN training computation framework, which only introduces the binary mask to guide the execution of matrix multiplications. And the bottom hardware---GPU only needs to support the basic matrix multiplication regardless of DNN model.

\section{Experiments}\label{sect:experiments}

\subsection{Experimental Methodology}

To evaluate the effectiveness of proposed approximate random dropout for MLP and LSTM, we compare it with the conventional dropout technique in terms of the accuracy and the training time. In section~\ref{ssect:dif_drop}, we vary different dropout possibility on a MLP to explore the influence of the dropout rate on the performance of a specific 4-layer MLP. Note that the dropout rate in our method refer to the target dropout rate $p$ as described in Section~\ref{ssect:sgd}.
In section~\ref{ssect:dif_network}, we compare different MLPs with a specific dropout rate.
The data set we use with MLP is MNIST.
LSTM is used in section~\ref{ssect:scaling_lstm} to verify the scalability of our method. The dataset we used with LSTM include a dictionary whose size is 8800, and the Penn Treebank~(PTB) data set which has long been a central data set for language modeling.
The experiment codes is implemented in Caffe and use a single GTX1080Ti GPU to run.

In addition, to evaluate the effectiveness of proposed dynamic pruning method for CNN, we use two widely used datasets---CIFAR-10 and MNIST. The CNN models we used are LeNet-5, AlexNet and VGG19, which cover a wide range of CNN with different parameter sizes. As for the baseline setting, we take the accuracy of the traditional CNN training of the selected model under the corresponding dataset as the baseline. In this section, we vary the prune possibility from 0.3 to 0.6 and record the accuracy and training time for each prune possibility.

\subsection{Results of Approximate Random Dropout}

\subsubsection{Comparison of different dropout rate}\label{ssect:dif_drop}

The structure of the 4-layer MLP is described as follow: the input layer is shaped according to the batch size; the output layer has 10 neurons for digit 0 to 9; the two hidden layers have 2048 neurons both. During training, we set the following hyper-parameters: the batch size is 128, the learning rate is 0.01, and momentum is 0.9.

\begin{figure}[!tb]
\centering
\includegraphics[width=0.7\linewidth]{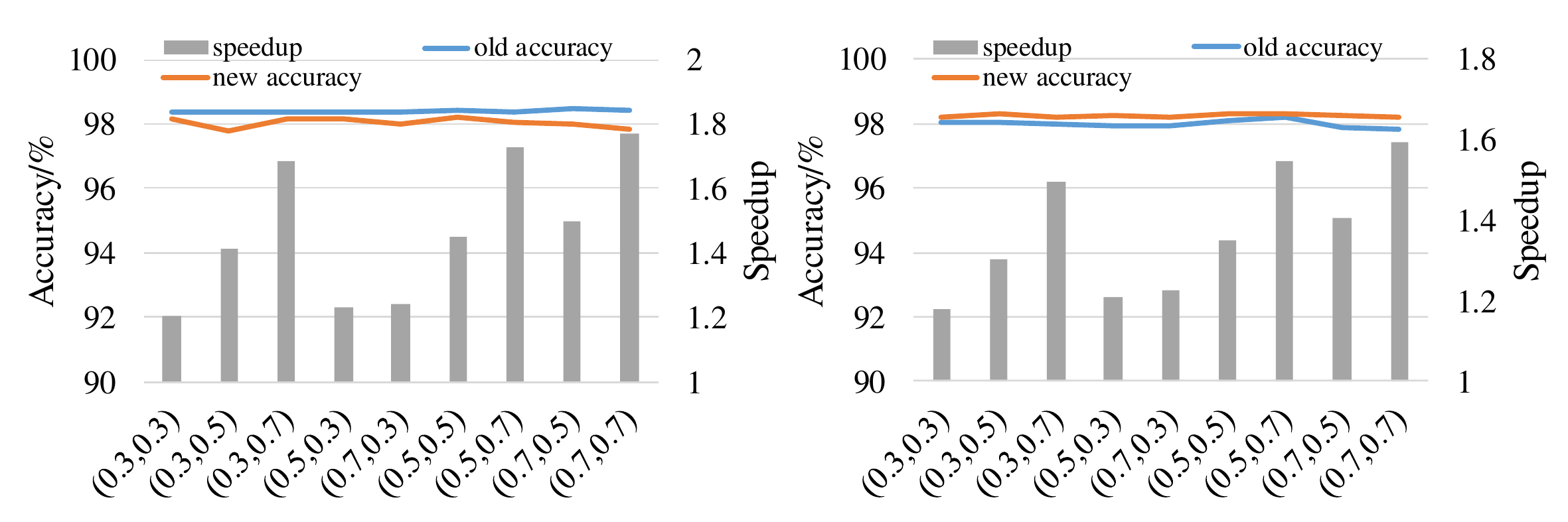} 
\caption{Comparing different dropout rate combinations on specific network.}
\label{dif_drop} 
\end{figure}

We vary the dropout rate from $(0.3, 0.3)$ to $(0.7,0.7)$ (two hidden layers may have varied dropout rate), and record the accuracy and training time for each dropout rate.
The comparison of two metrics of RDP and TDP against the conventional dropout are shown in Fig.~\ref{dif_drop}. The training time of conventional dropout is divided by the new training time of proposed approximate random dropout to obtain the speedup rate.

The results show that the speedup rate brought by RDP ranges from 1.20 to 1.77 compared with the traditional random dropout technique when the dropout rate varies between 0.3 and 0.7, which comply to our intuition as the amount of data that require no calculation expands with the increment of the dropout rate.
The speedup rate brought by TDP ranges from 1.18 to 1.6. The little slowdown is induced by the calculation of the nonzero positions in the output matrix before matrix multiplication.
The accuracy loss of these two classes of dropout patterns is less than $0.47\%$, which is the evadible concession to the speedup. TDP has less accuracy loss than RDP which can be attributed to the abundance of sub-models in TDP.

\subsubsection{Comparison of different networks}\label{ssect:dif_network}
We investigate the speedup in different MLP structures using a fixed dropout rate (0.7, 0.7).
Those MLPs have the same input and output layer as described in section~\ref{ssect:dif_drop}. Their hidden layer size is shown in Table~\ref{table_dif_network}. For instance, $1024\times64$ in the second column means the first and the second hidden layer's size are 1024 and 64, respectively.
The hyper-parameters of optimization algorithm follow above experiments.

From Table~\ref{table_dif_network}, the accuracy degradation is less than $0.5\%$. In some cases, the accuracy even increases.
Moreover, the speedup rate increases as the network size increases. Especially, in the case of $4096\times 4096$ network, both of the proposed dropout patterns reach a $2\times$ speedup.

\begin{table}[!tb]
\centering
\caption{Comparing different network with specific dropout rate}
\label{table_dif_network}
\begin{tabular}{|c|c|c|c|c|}
\hline
\multicolumn{1}{|l|}{\begin{tabular}[c]{@{}l@{}}Dropout \\ rate\end{tabular}} & \multicolumn{1}{l|}{\begin{tabular}[c]{@{}l@{}}Network \\ size\end{tabular}} & \multicolumn{1}{l|}{\begin{tabular}[c]{@{}l@{}}Dropout \\ pattern\end{tabular}} & \multicolumn{1}{l|}{Accuracy(its loss)} & \multicolumn{1}{l|}{\begin{tabular}[c]{@{}l@{}}Speedup \\ rate\end{tabular}} \\ \hline
\multirow{8}{*}{0.7}                                                           & \multirow{2}{*}{1024*64}                                                     & ROW                                                                             & 98.07\%(-0.42\%)                          & 1.27                                                                         \\ \cline{3-5}
                                                                               &                                                                              & TILE                                                                            & 98.11\%(-0.38\%)                          & 1.19                                                                         \\ \cline{2-5}
                                                                               & \multirow{2}{*}{1024*1024}                                                   & ROW                                                                             & 98.01\%(-0.35\%)                          & 1.45                                                                         \\ \cline{3-5}
                                                                               &                                                                              & TILE                                                                            & 98.15\%(-0.21\%)                          & 1.41                                                                         \\ \cline{2-5}
                                                                               & \multirow{2}{*}{2048*2048}                                                   & ROW                                                                             & 98.44\%(0.37\%)                           & 1.77                                                                         \\ \cline{3-5}
                                                                               &                                                                              & TILE                                                                            & 98.5\%(-0.31\%)                           & 1.60                                                                         \\ \cline{2-5}
                                                                               & \multirow{2}{*}{4096*4096}                                                   & ROW                                                                             & 98.00\%(-0.47\%)                          & 2.16                                                                         \\ \cline{3-5}
                                                                               &                                                                              & TILE                                                                            & 98.16\%(-0.31\%)                          & 1.95                                                                         \\ \hline
\end{tabular}
\end{table}

\subsubsection{Scaling to Long Short-Term Memory Model}\label{ssect:scaling_lstm}
We evaluate the speedup rate and the model performance on LSTM, which predicts the following word based on the given words. Each of the two hidden layers of LSTM contain 1500 neurons.
During training, we set the following hyper-parameters: the base learning rate is 1~(the base learning rate will gradually decrease), batch size is 20, the maximum epoch is 50, and the length of the sequence is 35.
Note that the execution of LSTM is also performed as matrix multiplication, thus our proposed approximate dropout can be easily applied to LSTM.

As shown in Table~\ref{table_large_network}, the accuracy degradation is less than $1\%$. When dropout rate is increasing, the speedup rate increases without undermining the accuracy loss.

\begin{table}[!tb]
\centering
\caption{A dictionary data set which contains 8800 words on LSTM.}
\label{table_large_network}
\begin{tabular}{|l|l|l|l|l|}
\hline
\multicolumn{2}{|l|}{dropout rate}   & (0.3,0.3) & (0.5,0.5) & (0.7,0.7) \\ \hline
\multirow{3}{*}{accuracy} & original & 47.9\%    & 47.3\%    & 45.9\%    \\ \cline{2-5}
                          & ROW      & 46.9\%    & 46.0\%    & 44.5\%    \\ \cline{2-5}
                          & TILE     & 47.2\%    & 46.5\%    & 44.4\%    \\ \hline
\multirow{3}{*}{speedup}  & original & 1.0       & 1.0       & 1.0       \\ \cline{2-5}
                          & ROW      & 1.18      & 1.47      & 1.53      \\ \cline{2-5}
                          & TILE     & 1.18      & 1.43      & 1.49      \\ \hline
\end{tabular}
\end{table}

\begin{figure}[!tb]
\centering 
\includegraphics[width=7cm]{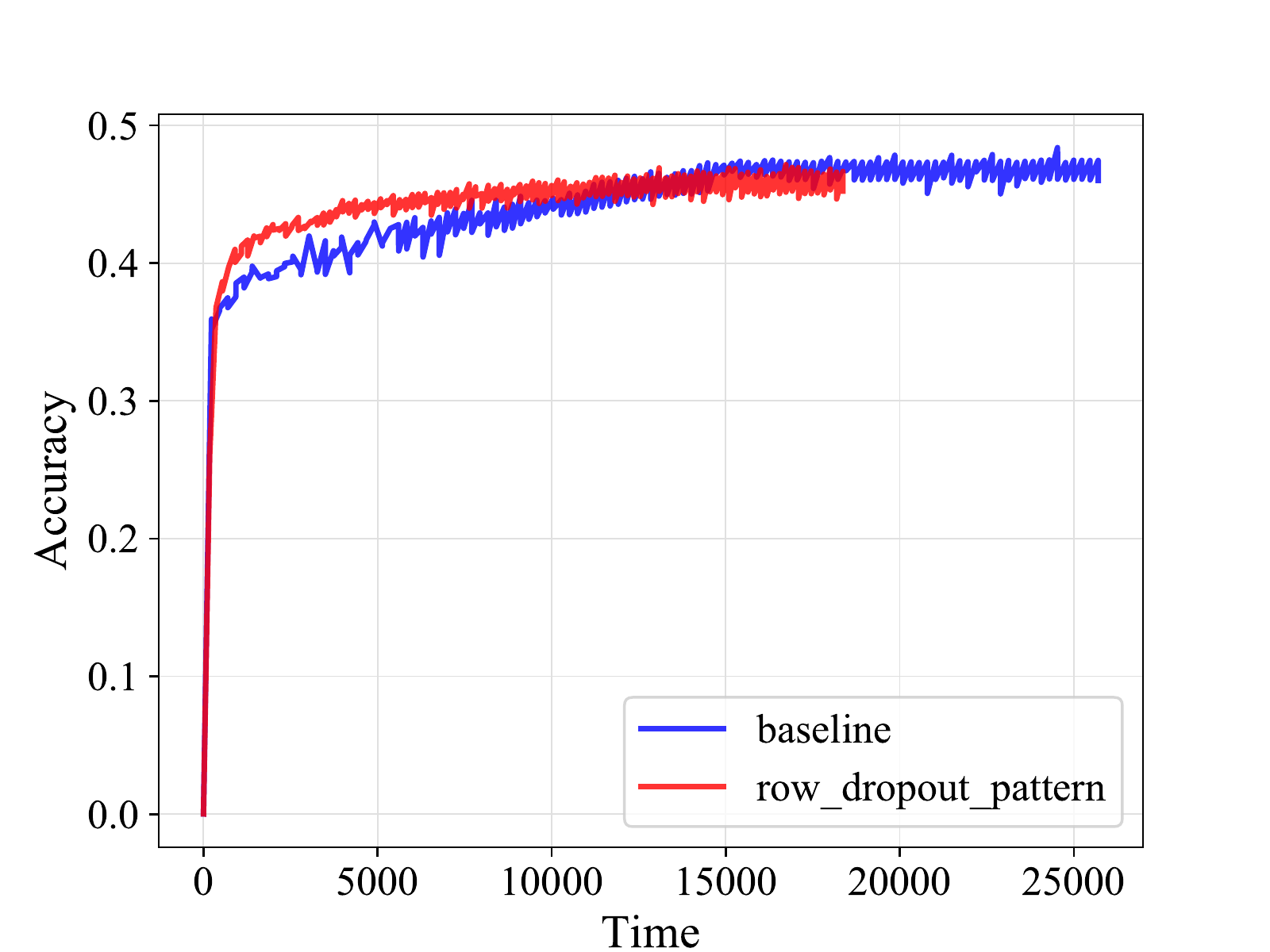} 
\caption{The training process of RDP and traditional random dropout.}
\label{train_log} 
\end{figure}

To illustrate the effectiveness of the proposed method, we fix the dropout rate to $0.5$ and trace the accuracy of RDP until it's convergence. As shown in Fig.~\ref{train_log}, the red curve records our approximate random dropout training process; the blue one records the traditional random dropout. The convergence of our method is earlier than the traditional random dropout. Moreover, the smoothness of red curve indicates the approximate random dropout is helpful for the training process.

\begin{figure}[!tb]\vspace{-5pt}
\centering 
\includegraphics[width=0.7\linewidth]{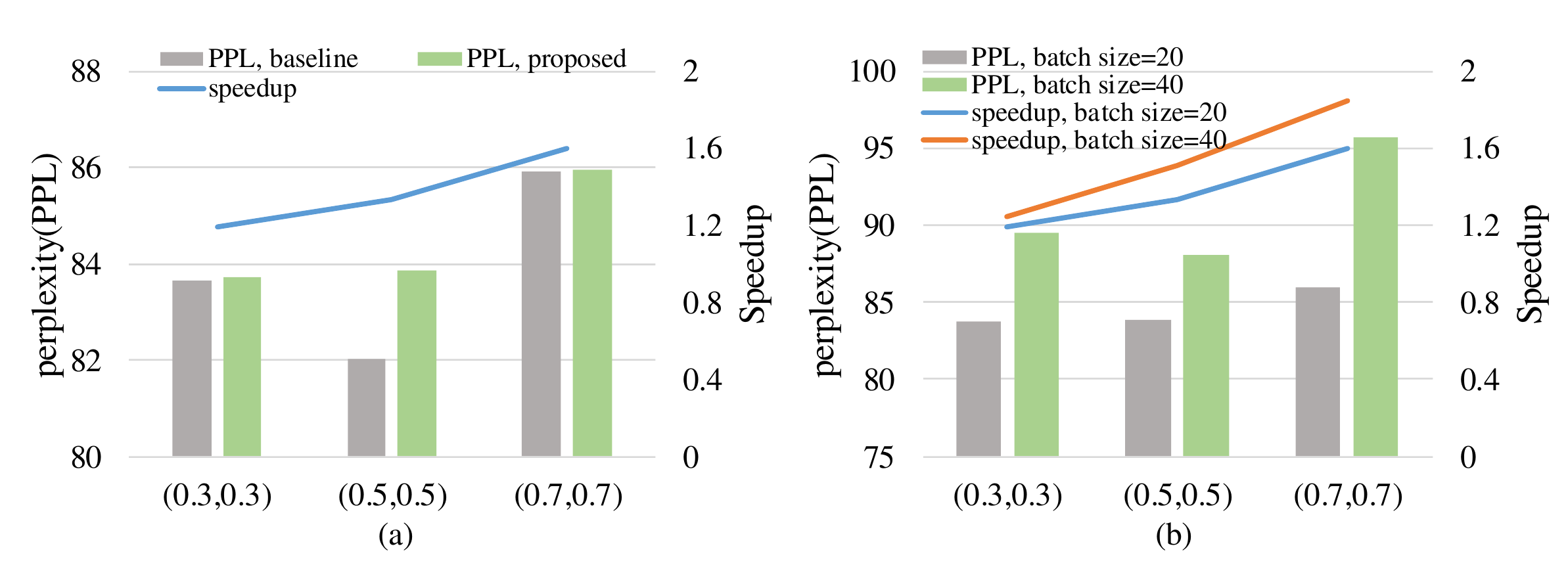} 
\caption{Speedup rate and accuracy of Row approximate dropout on 3-layer LSTM.}
\label{row_lstm} 
\end{figure}

The result using the Penn Treebank data set~(PTB) on the 3-layer LSTM is shown in Fig.~\ref{row_lstm}(a). The test perplexity using RDP only increases $0.04$ given the dropout rate is 0.7, which further shows that our proposed approximate dropout algorithm can generate adequate sub-models for PTB data set.
Besides, when dropout rate increases from 0.3 to 0.7, the speedup rate also increases from 1.2 to 1.6.

We vary the batch size from 20 to 40.
Noted that SGD based search and data initialization are an one-time effort. When the batch size is increased, only the matrix operation and data transmission time increase accordingly. As shown in Fig.~\ref{row_lstm}(b), the speedup rate increases when batch size increases.
However, since one dropout pattern is applied to the whole batch, the sub-models generated during training may not be sufficient, which raises the perplexity.

\subsection{Results of Dynamic Pruning Method}

\begin{figure}[!tb]
\centering
\includegraphics[width=0.5\linewidth]{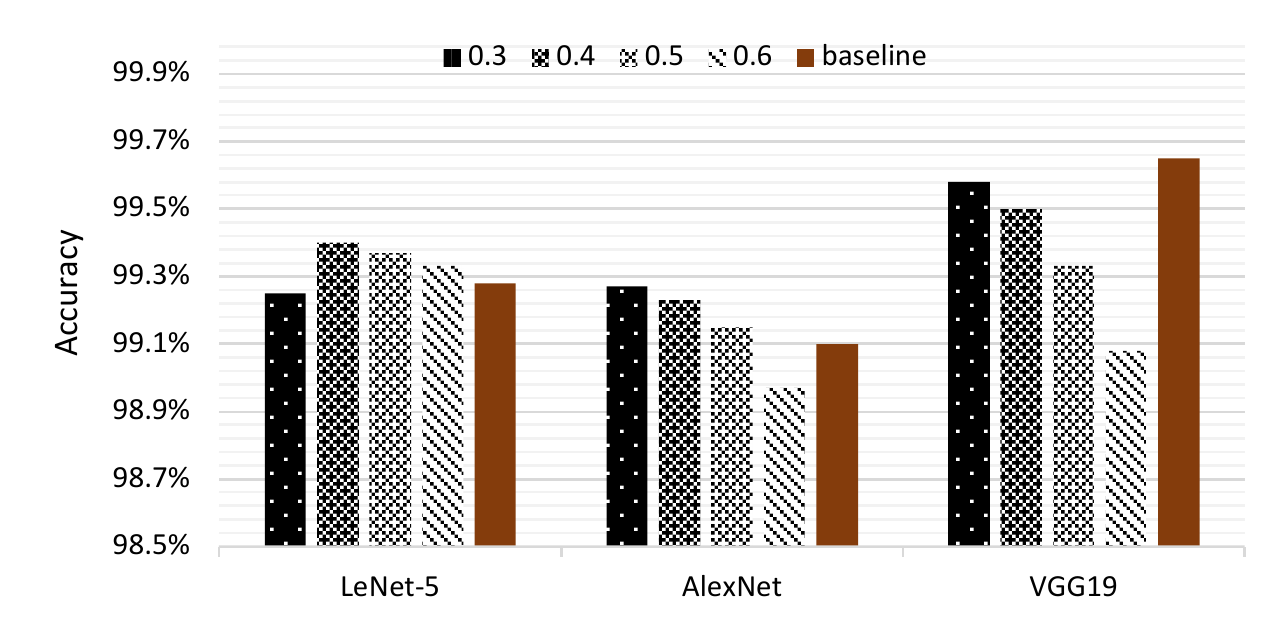} 
\caption{Accuracy after applying the RPP on MNIST dataset.}
\label{fig-row-mnist-acc}
\end{figure}

Fig~\ref{fig-row-mnist-acc} shows the CNN accuracy for three networks on MNIST dataset. For example, regarding VGG19, our dynamic pruning method shows negligible accuracy loss for MNIST compared to the baseline, even when the prune possibility is as high as $60\%$, accuracy loss can still be maintained at $0.57\%$. AlexNet and LeNet-5 perform even better and can guarantee the accuracy loss is less than $0.2\%$ when the prune possibility is $60\%$. Moreover, the CNN accuracy exceeds the baseline sometimes. This is because dynamic pruning method can reserve the key information in input feature map while reducing the redundant information during the training process.

\begin{figure}[!tb]
\centering
\includegraphics[width=0.5\linewidth]{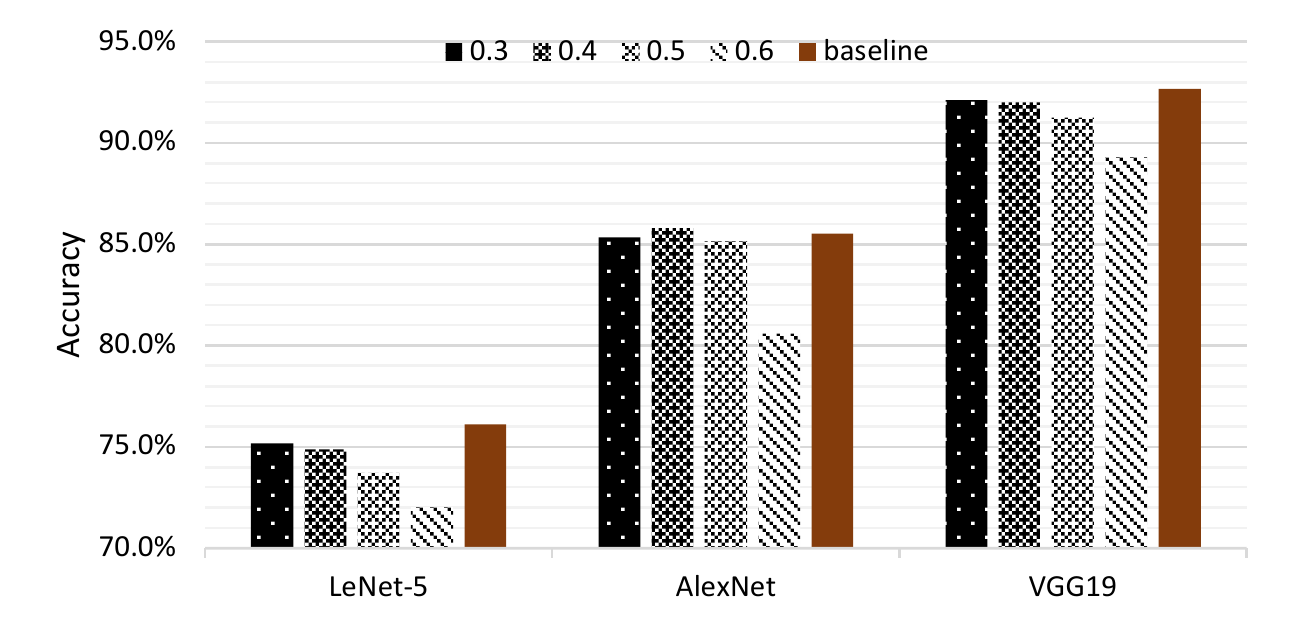} 
\caption{Accuracy after applying the RPP on CIFAR-10 dataset.}
\label{fig-row-cifar-acc}
\end{figure}

Fig.~\ref{fig-row-cifar-acc} shows the CNN accuracy for three networks on CIFAR-10 dataset by applying the RPP. Since CIFAR-10 dataset is more complex than MNIST dataset and its input feature map is more diverse, a larger prune possibility will lead to more accuracy loss. LeNet-5 can guarantee accuracy loss when the prune possibility is 0.4, while other two networks can maintain accuracy when the prune possibility is 0.5 as they are more redundant compared to LeNet-5. Therefore, for more complex datasets, we need to make a trade-off between prune possibility and accuracy loss.

\begin{figure}[!tb]
\centering
\includegraphics[width=0.5\linewidth]{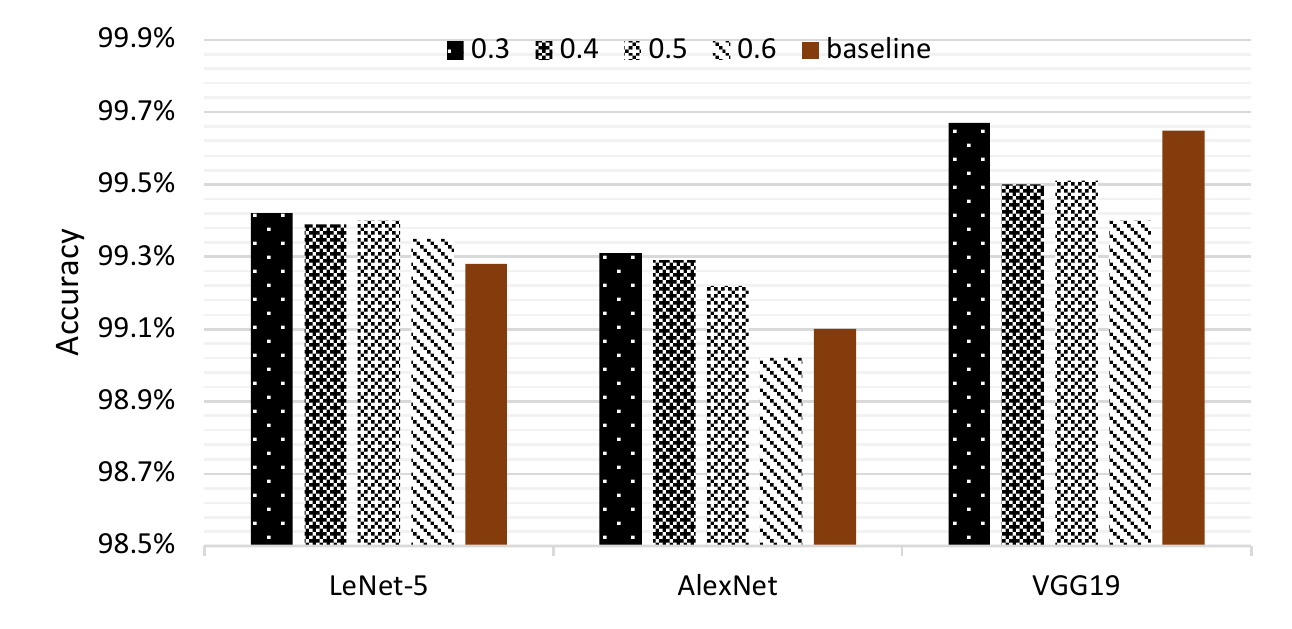} 
\caption{Accuracy after applying the BPP on MNIST dataset.}
\label{fig-block-mnist-acc}
\end{figure}

\begin{figure}[!tb]
\centering
\includegraphics[width=0.5\linewidth]{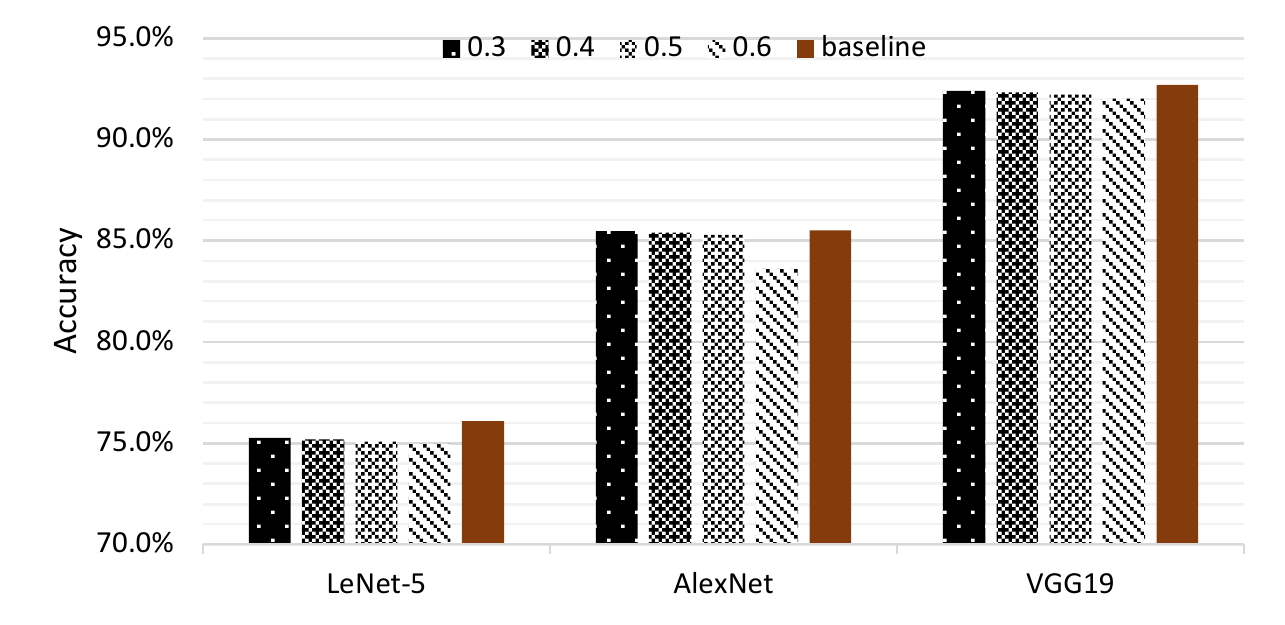} 
\caption{Accuracy after applying the BPP on CIFAR-10 dataset.}
\label{fig-block-cifar-acc}
\end{figure}

Fig~\ref{fig-block-mnist-acc} and Fig.~\ref{fig-block-cifar-acc} show the CNN accuracy for three networks on MNIST and CIFAR-10 datasets by applying the BPP. Compared to the RPP, the BPP can achieve higher accuracy when choosing the same pruning possibility. This is because the BPP can better match the sensitive region for satisfactory accuracy.

\begin{figure}[!tb]
\centering
\includegraphics[width=0.5\linewidth]{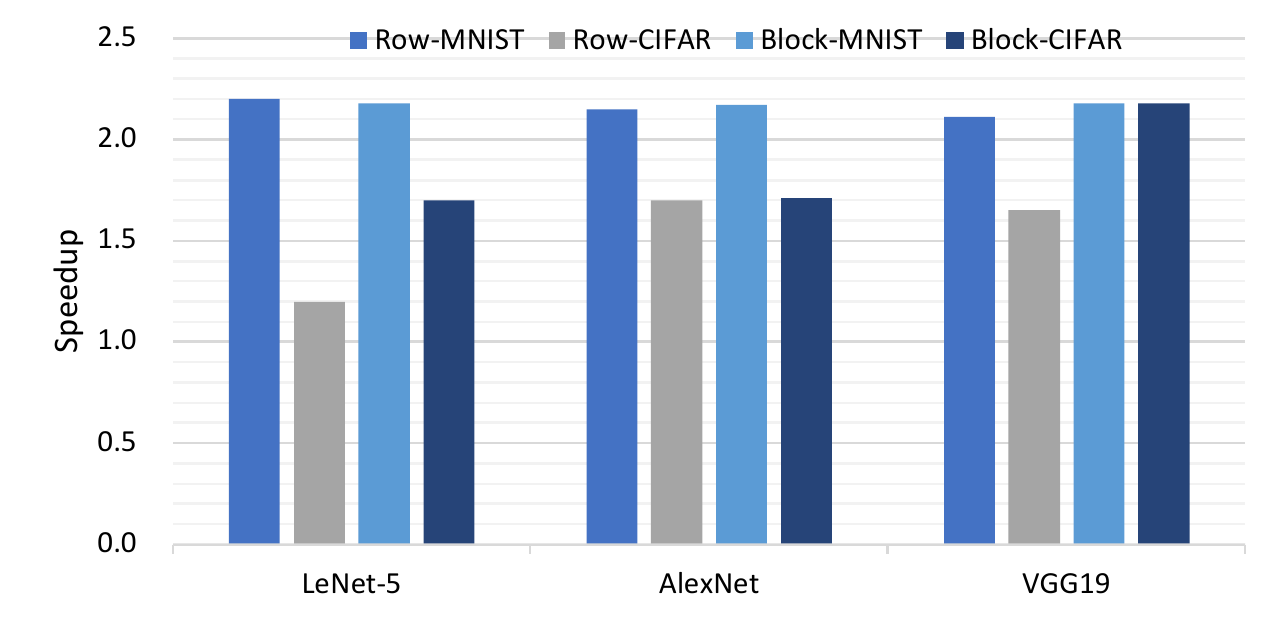} 
\caption{Speedup after applying the RPP and BPP on MNIST and CIFAR-10 datasets.}
\label{fig-conv-speedup}
\end{figure}

Fig.~\ref{fig-conv-speedup} shows the speedup for three networks. The speedup brought by RPP and BPP averagely achieves $2.15\times$ and $2.17\times$ on MNIST dataset, respectively. Moreover, since CIFAR-10 dataset is more complicated than MNIST dataset, the speedup brought by RPP and BPP only achieves $1.5\times$ and $1.9\times$ on CIFAR-10 dataset.

\section{Conclusion}\label{sect:conclusion}
Accelerating DNN training is nontrivial because it is difficult to leverage the sparsity of DNN in the dense matrix-multiplication. In this work, we first propose the approximate random dropout approach to eliminate the unnecessary multiplication and data access in MLP and LSTM by replacing the traditional random dropout with an approximate random dropout. The two classes of dropout patterns can avoid the divergence issue in GPU, reduce the scale of the matrix, and thus gain significant improvement on the energy-efficiency with marginal decline of the model performance. The proposed SGD-based search algorithm can guarantee the dropout rate of a single neuron or synapse is equivalent to the conventional random dropout, as well as the convergence and accuracy of the models. In addition, we propose the dynamic pruning method for accelerating CNN training process. The proposed method offers a promising opportunity to leverage the sensitivity in input feature maps to accelerate CNN training.
In general, the speedup rate ranges from 1.18-2.16~(1.24-1.85) when dropout rate is $0.3$-$0.7$ on MLP (LSTM) with negligible accuracy drop. And the speedup rate can reach to $2.17\times$ on CNN with satisfactory training accuracy.
The proposed method has been wrapped as an API and integrated into Caffe. The speedup can be much higher if the proposed method can be integrated into the cuBLAS Library.

\bibliographystyle{plain}
\bibliography{reference}

\end{document}